# 3-D Position Estimation from Inertial Sensing: Minimizing the Error from the Process of Double Integration of Accelerations

Pedro Neto and J. Norberto Pires
Department of Mechanical Engineering, CEMUC
University of Coimbra
POLO II, 3030-788, Coimbra, Portugal
{pedro.neto,norberto}@dem.uc.pt

Anónio Paulo Moreira
ROBIS - INESCTEC Porto
University of Porto
Rua Dr. Roberto Frias, 4200-465, Porto, Portugal
{amoreira}@fe.up.pt

*Abstract*—This paper introduces a new approach to 3-D position estimation from acceleration data, i.e., a 3-D motion tracking system having a small size and low-cost magnetic and inertial measurement unit (MIMU) composed by both a digital compass and a gyroscope as interaction technology. A major challenge is to minimize the error caused by the process of double integration of accelerations due to motion (these ones have to be separated from the accelerations due to gravity). Owing to drift error, position estimation cannot be performed with adequate accuracy for periods longer than few seconds. For this reason, we propose a method to detect motion stops and only integrate accelerations in moments of effective hand motion during the demonstration process. The proposed system is validated and evaluated with experiments reporting a common daily life pick-and-place task.

## I. Introduction

The reliability, size, intrusive character and cost of some existing motion tracking technologies have hindered the development of some areas of robotics. This is particularly true, for example, in areas related with robot autonomy and programming by demonstration. Each interaction technology has advantages and disadvantages. Some hybrid systems combining different types of technology (inertial, magnetic, optical and GPS based technologies) have shown good results. Nevertheless, reliable and accessible 3-D position estimation is still a problem.

### A. Interaction Technologies and methods

Interaction technologies for human motion tracking have increasingly being subject of study by researchers. A review on motion capture technologies and current challenges associated to their application in robotic systems is presented in [1]. Different methods have been employed to approach problems related to modeling and recognition of human behaviors [2] and motion tracking. Basic concepts for mapping typical human actions performed in a household environment to a robotic system are explained in [3]. Magnetic and optical sensors allow obtaining an absolute reference for the system in study and do not suffer from the problem of drift that the inertial sensors suffer. A major drawback of magnetic-based sensors is its sensitivity to magnetic distortions in the Earth's magnetic field. Ekvall and Kragic explore grasp recognition in a PbD system using a magnetic tracker to capture motion [4]. Sugiyama and Miura develop a vision-based interface in which the user can instruct a robot by making it move in the same way as the user's motion [5]. These vision-based systems present some important drawbacks such as the limited fields of view, occlusions, etc.

Inertial-based motion capture systems rely on acceleration and rotational velocity measurements from accelerometers and gyroscopes, respectively. Inertial tracking suffers from severe drift problems: high noise and large uncertainties such as bias and scale factor. Thus, they cannot provide accurate pose information during continuous operation (long term stability is affected). Despite the inherent problems associated with these sensors, the overall performance can be improved by combining them with other sensors. There are many possibilities to combine individual sensors into a new multi-sensorial system. The positive aspects of different sensors can be explored and combined, originating a "better" sensor. The small sensors that combine inertial and magnetic sensing are usually called miniature magnetic and inertial measurement units (MIMUs). Maeda *et al.* present a sensor-based system that measures full body motion of the user [6]. Miller *et al.* report the use of a set of inertial sensors to control the robot arm of NASA Robonaut [7]. A real-time hybrid solution to articulated 3D arm motion tracking for home-based rehabilitation by combining visual and inertial sensors is presented in [8]. An inertial-sensor-based hybrid tracking technology is presented in [9]. Sessa *et al.* present a method for reliable comparison among IMUs using a Vicon system as reference measurement system [10]. Ojeda and Borenstein present a navigation system for walking persons based on a 6-axis IMU attached to the user's boot [11]. They propose a technique known as "Zero Velocity Update" that virtually eliminates the ill-effects of drift in the accelerometers. Another study presents a 3-D position tracking system composed by an IMU and an external marker-based video tracking solution [12]. Drift is reduced by fusing measurements from both sensors using an extended Kalman filter.

## B. Proposed Approach

This paper presents a new approach to 3-D position and velocity estimation from acceleration data, i.e., a 3-D motion tracking system having a small size and low-cost MIMU composed by both a digital compass and a gyroscope as interaction technology, Fig. 1. By combining inertial and magnetic sensing we ensure not only fast motion tracking but also relatively long term stability. Hand positions are estimated by double integrating motion data (accelerations due to effective hand motion) provided by a 3-axis accelerometer embedded into the digital compass, Fig. 2. Since the accelerometer provides both accelerations due to motion and gravity, it is proposed a method to separate these two acceleration components. Roll and pitch angles are acquired from the digital compass and yaw angles are estimated by fusing data from the digital compass and the gyroscope using the Kalman filter (this process is out of the scope of this paper).

A major challenge is to minimize the error caused by the process of double integration of accelerations. Owing to drift error (error accumulates with time), position estimation cannot be performed with an adequate accuracy for periods longer than few seconds. For this reason, we propose a method to detect hand motion stops and only integrate accelerations in effective moments of hand motion during the demonstration process. The proposed system is validated and evaluated with experiments reporting a common daily life pick-and-place task. Errors are analyzed and discussed.

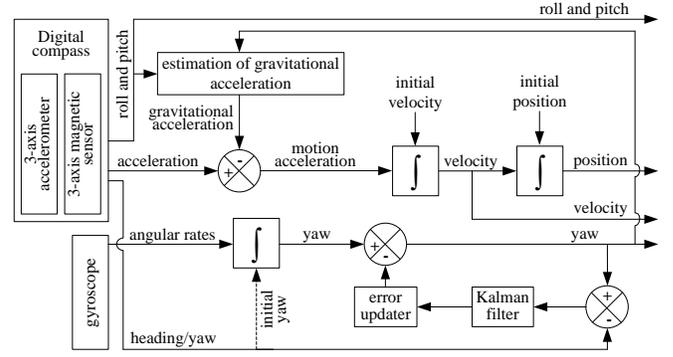

Fig. 2. Block diagram of the proposed system.

## II. POSITION AND VELOCITY ESTIMATION

For the sake of clarity, it is important to establish at this stage the reference frames involved in the system, Fig. 3:

*1) Inertial frame {I}: Inertial sensors provide measurements relative to an inertial frame. The axes $x_I$ and $y_I$ are located in the local level tangent plane, while the $z_I$ axis points in the opposite direction to the center of the Earth.*

*2) Body frame {B}: This frame is attached to the MIMU in/with a desired location and orientation.*

*3) Navigation frame {N}: The navigation frame is attached to a fixed point on the surface of the Earth.*

In order to convert hand motion into robot motion, the measurements made on the inertial and body frame must be transformed (mapped) to the navigation frame by using a proper rotation matrix and translation vector defining the initial MIMU pose in relation to the navigation frame. These ones are made known to the robot by means of the calibration process.

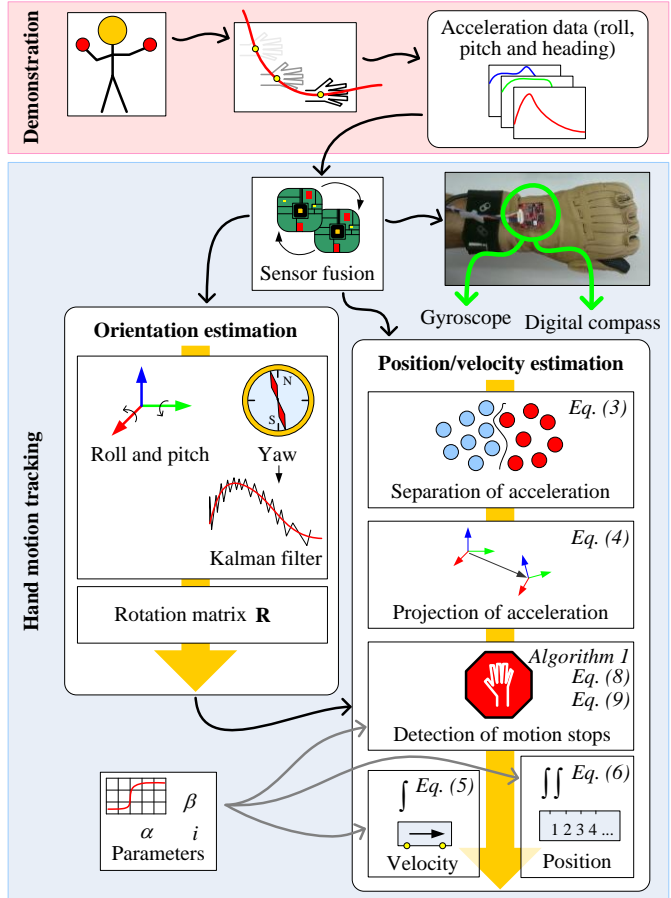

Fig. 1. Layout of the proposed system.

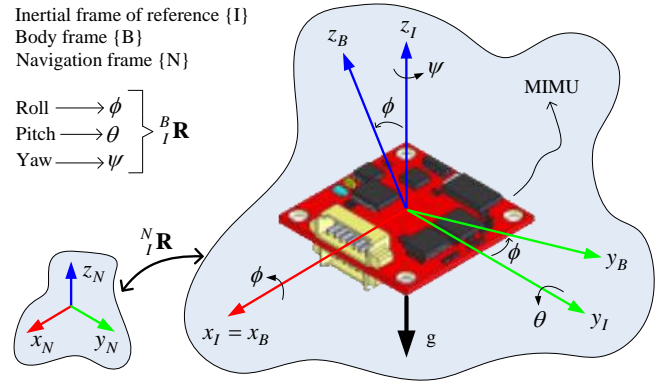

Fig. 3. Reference frames.

Accelerations can be mathematically integrated once to obtain velocity and twice to obtain changes in position. Only the accelerations due to effective hand motion have to be integrated. Since the accelerometer provides a combination of accelerations due to motion, gravity and error, it becomes necessary to separate these acceleration components.

Errors in measured accelerations can be deterministic (bias, scale factor and axis misalignment) and/or random errors. Such errors propagate through the integration process, causing a considerable drift in estimated positions. Owing to drift error, position estimation cannot be performed with adequate accuracy for periods longer than few seconds [13-14]. For this reason, we are proposing a method to only integrate accelerations in effective moments of hand motion during the demonstration process. In addition, since we are using estimated orientations to separate and project accelerations, errors in estimated roll, pitch and yaw angles also affect position estimation accuracy. A small error in estimated orientation can yield a large error in estimated position.

### A. Sepparation of Acceleration

The accelerations $\mathbf{a} = (a_x, a_y, a_z)^T$ given by the 3-axis accelerometer are a combination of accelerations due to motion $\mathbf{a}_m$, a gravitational component $\mathbf{a}_g$ and error $\boldsymbol{\varepsilon}$:

$$\mathbf{a} = \mathbf{a}_m + \mathbf{a}_g + \boldsymbol{\varepsilon} \quad (1)$$

Since only $\mathbf{a}_m$ is integrated to estimate positions/displacements, it becomes necessary to separate $\mathbf{a}$ into its component parts highlighted in (1).

The accelerations $\mathbf{a}$ are given in the body frame {B}. At the same time, the gravity vector $\mathbf{g} = (0, 0, g)^T$ has a constant value and direction, being $g$ the Earth's gravity. $\mathbf{g}$ is expressed in the inertial frame {I}, Fig. 3. Knowing the orientation of the MIMU in each instant of time we can estimate the acceleration $\mathbf{a}_g$ projected in {B} by recurring to $\mathbf{g}$ and to a rotation matrix describing the transformation from frame {I} to {B}, $^B_I\mathbf{R}$. This matrix is defined with estimated roll, pitch and yaw angles from the MIMU in each instant of time. Thus, $^B_I\mathbf{R}$ performs the coordinate transformation of $\mathbf{g}$ into a vector in body frame {B}:

$$\mathbf{a}_{g,B} = {}^B_I\mathbf{R}\,\mathbf{g} \quad (2)$$

From (1), if the MIMU is static we have that $\mathbf{a} = \mathbf{a}_g + \boldsymbol{\varepsilon}$. On contrary, if the MIMU is moving, from (1) and (2), the acceleration due to motion in frame {B} can be computed by the following:

$$\mathbf{a}_{m,B} = \mathbf{a} - \left({}^B_I\mathbf{R}\,\mathbf{g}\right) - \boldsymbol{\varepsilon} \quad (3)$$

It is assumed that $\boldsymbol{\varepsilon} = 0$. Owing to the constantly changing orientation of the MIMU during a demonstration, real-time updates of estimated orientations are required to define $^B_I\mathbf{R}$.

### B. Projection of Acceleration

Accelerations due to motion defined in relation to {B}, Fig. 3, must be projected into a navigation frame {N} defined *a priori* (the frame that the "robot knows" from the calibration process). As we know $^B_I\mathbf{R}$ and $^N_I\mathbf{R}$, and $^I_N\mathbf{R} = {}^N_I\mathbf{R}^T$, we have that $^B_N\mathbf{R} = {}^B_I\mathbf{R}\,{}^I_N\mathbf{R}$. Finally, the accelerations due to motion expressed in {N} are:

$$\mathbf{a}_{m,N} = {}^B_N\mathbf{R}^T\,\mathbf{a}_{m,B} \quad (4)$$

From (4) it can be clearly seen that error in estimated orientation angles promotes error in projected accelerations.

### C. Numerical Integration of Acceleration

The MIMU velocities $\mathbf{v}(n) = (v_x, v_y, v_z)^T_n$ at an instant of time $t_n$ can be estimated by accumulating velocity changes:

$$\mathbf{v}(n) = \mathbf{v}(0) + \sum_{k=1}^{n}\mathbf{a}_{m,N}(k)\,\Delta t \quad (5)$$

In which $\mathbf{v}(0)$ is the initial velocity and $\Delta t$ the integration time. The displacements $\mathbf{s}(n) = (s_x, s_y, s_z)^T_n$, can be estimated by accumulating changes in position:

$$\mathbf{s}(n) = \mathbf{s}(0) + \sum_{k=1}^{n}\mathbf{v}(k)\,\Delta t \quad (6)$$

### D. Detection of Motion Stops

During the demonstration of a given task there are moments in which the demonstrator hand (with the MIMU attached) is in effective motion and others in which the hand is stopped. It becomes necessary to identify these moments to only integrate motion accelerations corresponding to the moments of hand motion. First, we need to compute the resulting acceleration due to motion for each instant of time:

$$a_{mr} = \sqrt{a_{mx}^2 + a_{my}^2 + a_{mz}^2} \quad (7)$$

In which $a_{mx}$, $a_{my}$ and $a_{mz}$ are the components of $\mathbf{a}_{m,N}$. To detect motion stops we propose to compute the variance of each value of resulting acceleration:

$$\sigma^2(i) = \frac{1}{n-1}\sum_{j=i-n+1}^{i}\left(a_{mr}(j) - \bar{a}_{mr}(j)\right)^2 \quad (8)$$

Being $n$ the population size and $\bar{a}$ the population (accelerations) average. For this specific problem the population average can be defined by:

$$\bar{a}_{mr}(i) = \frac{1}{n}\sum_{j=1}^{n}a_{mr}(j) \quad (9)$$

The moments corresponding to hand motion are defined by establishing a threshold value $\lambda_v$ applied to the computed variances (8) and Alghorithm 1. The value of $\lambda_v$ is defined by trial-and-error and it directly influences error estimation (it is defined according to the sensibility of the accelerometer).

## III. EXPERIMENTS

Having worn the data glove and with the MIMU (*OceanServer OS500-US*) attached to the hand, the demonstrator performs a demonstration that *a posteriori* is supposed to be repeated by a robot. The demonstration task consists in a pick-and-place operation, more specifically the transportation of a plastic bottle from one location to another for the purpose of putting water in a plastic cup. This is a daily life task that presents a clear success/failure criterion.

---

**Algorithm 1** Detection of hand motion

**Input:** variance $\sigma^2$, threshold for variance $\lambda_v$,
threshold for motion detection $\lambda_m$ (lenght of motion)

**Output:** motion detector *m*, begining of motion space *b*,
end of motion space *e*, number of readings *n*

```
1:  Begin
2:    n ← 0
3:    For Each σ² Do
4:      n ← n + 1
5:      If (σ² ≥ λv) Then
6:        m(i) ← 1
7:      Else
8:        m(i) ← 0
9:      End If
10:   End For
11:   For i = 1 To n
12:     /* definition of the begining of motion space b */
13:     If (m(i) = 1) Then
14:       VarB = true
15:       b = i
16:       While (VarB = true) Do
17:         For j = 1 To λm
18:           If ( m(i + j) = 1) Then
19:             updated ← i + j
20:           End If
21:         End For
22:         If (updated > i) Then
23:           VarB = true
24:           i ← updated
25:           /* definition of the end of motion space e */
26:         Else
27:           VarB = false
28:           e ← updated
29:         End If
30:       End While
31:     End If
32:   End For
33: End
```

### A. Tests

Two different experimental tests for the same pick-and-place task are presented. In *Test 1* the user demonstrates the task respecting the following actions, Fig. 4:

- Grasp the plastic bottle, Fig. 4 (a).

- Move the plastic bottle in vertical direction along *z* axis, Fig. 4 (b), and stop for a while.

- Move the bottle along *x* axis, Fig. 4 (c), and stop for a while.

- Move the bottle in direction to the plastic cup (essentially along *y* axis), Fig. 4 (d), and stop for a while, Fig. 4 (e).

- The hand rotates as if the demonstrator were to put water into the plastic cup, Fig. 4 (f). The final target is the centre of the plastic cup.

*Test 2* differs from *Test 1* in the way the demonstrator performs the task. In this case the demonstrator transports the plastic bottle directly to the target, without motion stops.

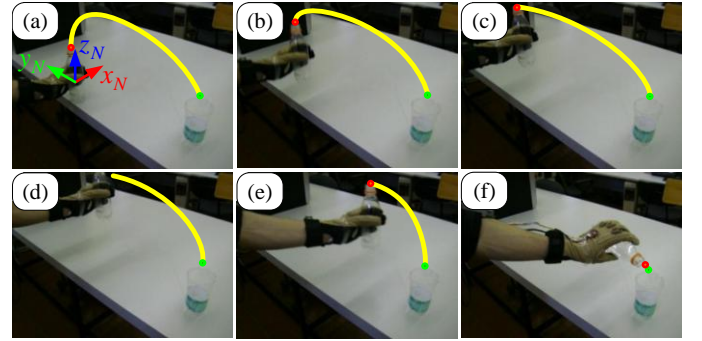

Fig. 4. Demonstration actions for *Test 1*.

### B. Results and Discussion

Fig. 5 shows the resulting acceleration due to motion established in (7) for *Test 1*.

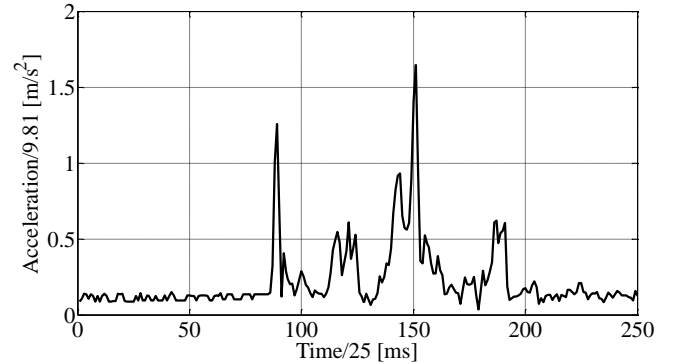

Fig. 5. Resulting accelerations due to motion for *Test 1*.

There follows the computation of the variance of the resulting accelerations (considering that *n*=2 in (8)), Fig. 6. To detect motion stops during the demonstration, the Alghorithm 1 is applied with a threshold value $\lambda_v = 0.01$. The result shown in Fig. 7 clearly indicates the moments in which the demonstrator hand is moving and stopped. These moments are highlighted in Table I. It presents details about the interval of the demonstration in which each different action occurs and how much long it takes.

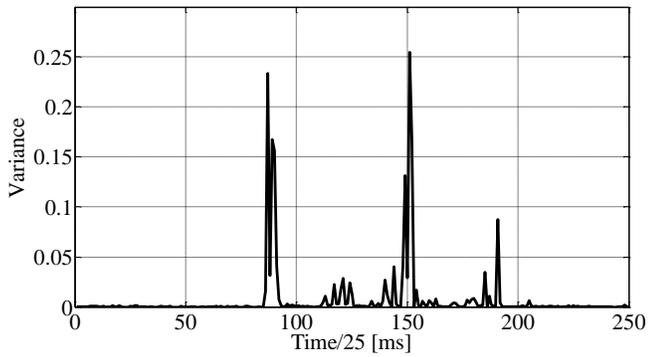

Fig. 6. Variance of resulting accelerations due to motion for *Test 1*.

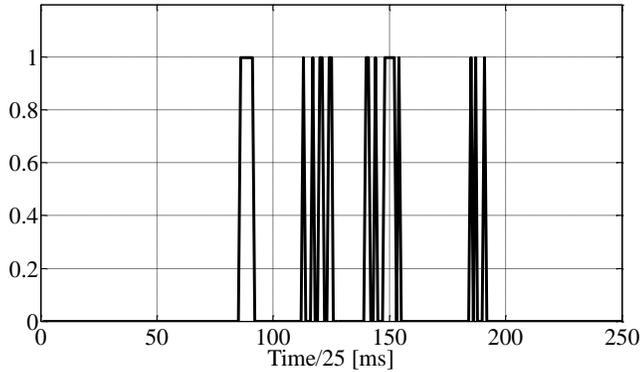

Fig. 7. Moments of hand motion stops (value 0) for *Test 1*.

TABLE I. DIFFERENT ACTIONS OF THE DEMONSTRATION (*TEST 1*)

| Interval of time | Description | Time [ms] |
|---|---|---|
| 0 - 86 | No hand motion occurs | 2150 |
| 87 - 94 | Vertical motion (mainly along *z* axis) | 175 |
| 95 - 114 | No hand motion occurs | 475 |
| 115 - 126 | Horizontal motion (mainly along *x* axis) | 275 |
| 127 - 140 | No hand motion occurs | 325 |
| 141 - 155 | Horizontal motion (mainly along *y* axis) | 350 |
| 156 - 185 | No hand motion occurs | 725 |
| 186 - 192 | Hand rotation | 125 |
| 193 - 250 | No hand motion occurs | 1425 |

The estimated hand path poses (positions and orientations) are represented in a 3-D graph, Fig. 8. The hand orientation is represented by a set of three vectors corresponding to each column of the rotation matrix defined by estimated roll, pitch and yaw angles. The projected views of the 3-D graph help to visualize the hand path, especially when such views are complemented with schematic drawings of the plastic bottle and cup. The view *yx* is shown in Fig. 9 and the view *yz* in Fig. 10.

For *Test 2* the estimated path poses are represented in Fig. 11 (view *yx*) and in Fig. 12 (view *yz*). Error is difficult to estimate because we do not have a nominal path to compare with the estimated one. On the other hand, we have a well defined target, the centre of the plastic cup.

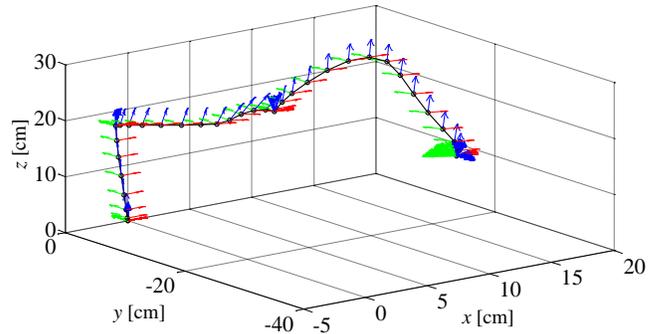

Fig. 8. Estimated hand path poses for *Test 1*.

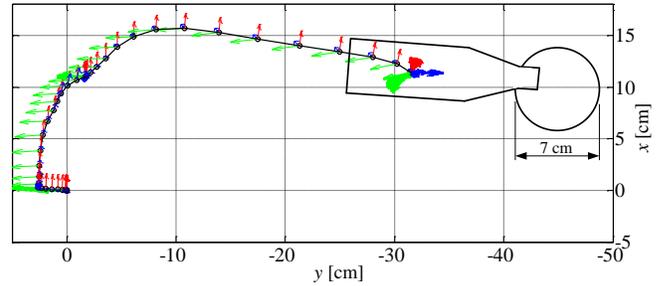

Fig. 9. Estimated hand path poses for *Test 1*, view *yx*.

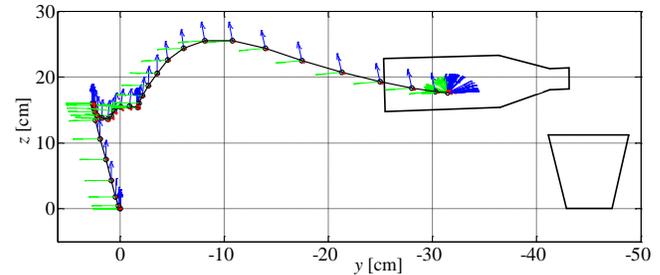

Fig. 10. Estimated hand path poses for *Test 1*, view *yz*.

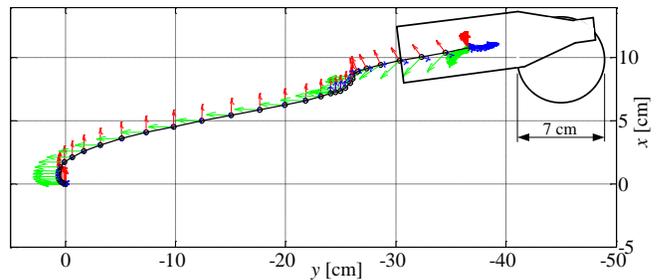

Fig. 11. Estimated hand path poses for *Test 2*, view *yx*.

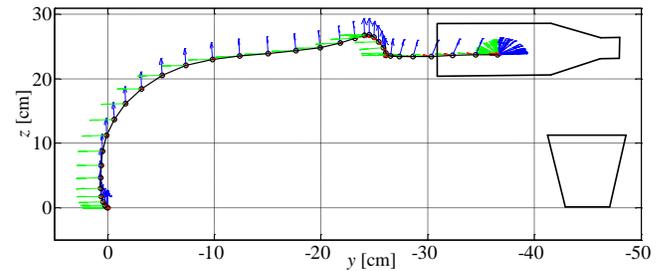

Fig. 12. Estimated hand path poses for *Test 2*, view *yz*.

Analyzing Fig. 9, Fig. 10, Fig. 11, Fig. 12, it can be stated that the error in plane *xy* for *Test 1* is about 1 cm along *x* axis and 2 cm along the *y* axis. For *Test 2* the error is about 2 cm along *x* axis and 2 cm along the *y* axis. In relation to the error along the *z* axis, analyzing the recording of both demonstrations it can be concluded that for both tests the error is about less than 1 cm. In reality, when we are dealing with magnetic and inertial sensing we cannot have an accurate estimate to error because it depends on several factors:

- Propagation of the error from estimated roll, pitch and yaw angles (definition of rotation matrix **R**).
- Double integration of accelerations (drift).
- Error associated with the hardware itself.
- Environmental factors (temperature, etc.).

All these factors result in a rapidly accumulating error in estimated positions. Some researchers have studied the way the error evolves when positions are obtained by double integrating accelerations. An interesting study in the field reports that in specific conditions error increases as $t^{1.5}$, being *t* the integration time [15]. Thong *et al.* study the error dependence on accelerometer noise [16]. Woodman analyses in detail some of the most important issues related with inertial navigation systems, including the error behavior [14]. Nevertheless, there is a lack of consensus regarding how errors behave with integration time. As error increases with time, the proposed solution to only integrate accelerations in moments of effective motion revealed to be a good option.

Analyzing the experimental tests from a practical perspective we can state that in *Test 1* the demonstrated task can be realized with success so that the estimated hand poses are transferred to a robot that executes the task, Fig. 13. In *Test 2* the water would be "placed" outside the plastic cup [17].

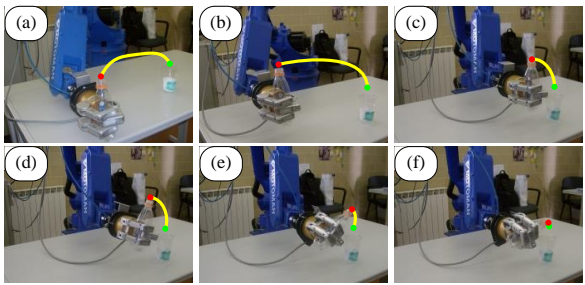

Fig. 13. A robot reproducing the demonstrated task in *Test 1*.

## IV. CONCLUSIONS AND FUTURE WORK

A new method for estimating 3-D positions from acceleration data has been described. It is based on the double integration of accelerations due to motion from a small size and low-cost MIMU composed by both a digital compass and a gyroscope. It was presented a method to separate accelerations due to motion from accelerations due to gravity, and another one to detect motion stops during a demonstration. Error in position estimation exists. It comes from different sources such as the MIMU hardware and from the process of double integration of accelerations. These increasing errors from the process of double integration of accelerations can be reduced if accelerations are integrated in short intervals of time, when effective motion occurs. Experiments demonstrated that the achieved error is still too large (especially for relatively long periods of time) to consider the proposed solution a reliable alternative to existing ones, namely the magnetic and vision based tracking systems. Nevertheless, the tolerance to magnetic fields, occlusions and the low-cost nature of the proposed system makes it a promising motion tracking solution.

Future work will focus on reducing the error associated to position estimation by improving the hardware that composes the MIMU. The system performance has to be validated with more tests performed by different people.

ACKNOWLEDGMENT

This research was supported by the Portuguese Foundation for Science and technology, PTDC/EME-CRO/114595/2009.